# PIRVS: An Advanced Visual-Inertial SLAM System with Flexible Sensor Fusion and Hardware Co-Design

Zhe Zhang, Shaoshan Liu, Grace Tsai, Hongbing Hu, Chen-Chi Chu, and Feng Zheng

*Abstract*— In this paper, we present the PerceptIn Robotics Vision System (PIRVS) system, a visual-inertial computing hardware with embedded simultaneous localization and mapping (SLAM) algorithm. The PIRVS hardware is equipped with a multi-core processor, a global-shutter stereo camera, and an IMU with precise hardware synchronization. The PIRVS software features a novel and flexible sensor fusion approach to not only tightly integrate visual measurements with inertial measurements and also to loosely couple with additional sensor modalities. It runs in real-time on both PC and the PIRVS hardware. We perform a thorough evaluation of the proposed system using multiple public visual-inertial datasets. Experimental results demonstrate that our system reaches comparable accuracy of state-of-the-art visual-inertial algorithms on PC, while being more efficient on the PIRVS hardware.

## I. INTRODUCTION AND RELATED WORK

### A. SLAM Background

SLAM has been an active topic for many years [1] because it provides two fundamental components for many applications: where I am, and what I see. While the theories for SLAM matured over the years, the challenge remains for SLAM to adapt to real-world applications. Each application has its own challenges. For example, a mobile robot requires mapping and localization in a large-scale environment, such as an entire building, which poses challenges for loop closing and large-scale optimization. On the other hand, augmented reality (AR) and virtual Reality (VR) applications require high-precision, jitter-free position tracking with low-latency to provide immersive user-experiences when viewing the virtual contents. For autonomous vehicles, localization using multiple sensors to handle various environment is essential, which poses challenges for real-time sensor fusion for tracking. Thus, the challenges for SLAM to work in real-world applications become the choice of sensors, the design of the system for the targeting application, and the implementation details.

### B. SLAM Approaches

SLAM approaches evolve with the development of sensors and computation platforms. At first, SLAM is mostly applied on robots equipped with wheel encoders and range sensors. Such SLAM system uses a Kalman Filter [2] with the assumption of linearly approximated model with Gaussian noise to jointly estimate the robot pose and a map (e.g., a set of landmarks), or a particle filter [3] to build multiple hypotheses to localize within a global map. Among all range sensors, laser range finder were especially popular because of its accuracy and stability. For example, in the 2005 DARPA Grand Challenge [4], five Sick AG LIDARs were mounted on the roof with a GPS for localization and mapping. Due to hardware limitations, vision-based SLAM algorithms were not popular at that time.

Figure 1. PIRVS Hardware (product name: Ironsides).

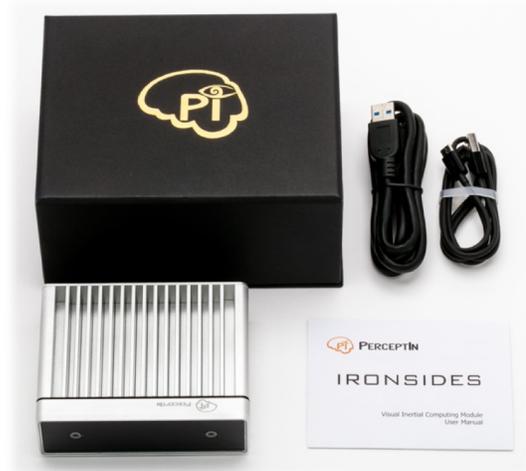

In 2009, Kinect [5], a low-cost active depth sensor, was introduced, which opened up new approaches for SLAM with 6-dof pose and 3D dense map (e.g., KinectFusion [6], ElasticFusion [7]). At the same time, passive sensors, like CCD/CMOS sensors, were becoming cost effective, better in quality, and ready for mass production, and therefore, many visual SLAM approaches using sparse features (e.g., PTAM [8], ORB-SLAM [9]) were developed. While most visual SLAM approaches build a sparse point cloud with image features, some use image-to-image alignment to directly estimate pose and depth (e.g., LSD-SLAM [10]), which could produce a denser map.

During the last decade, smart mobile devices, such as iPhone and Android phones, provide a whole new computation platform with low-cost MEMS-based inertial measurement unit (IMU) and a low-cost camera for SLAM. Many visual-inertial SLAM approaches [11-17] have been proposed and have been popular ever since. These approaches are commonly known as vision-aided inertial navigation system (VINS) or visual-inertial odometry (VIO). The combination is especially charming because the IMU provides high-frame rate pose prediction while cameras provide accurate pose correction and

Zhe Zhang, Shaoshan Liu, Grace Tsai, Hongbing Hu, Chen-Chi Chu, and Feng Zheng are with PerceptIn, Inc., Santa Clara, CA 95054, USA (e-mail: {zhe.zhang, shaoshan.liu, grace.tsai, hongbing.hu, jason.chu, feng.zheng} @perceptin.io).

map construction. There are two categories of visual-inertial systems: tightly-coupled [12-17] and loosely-coupled [11]. A tightly-coupled system jointly optimizes over both inertial and visual sensor measurements, which provides higher accuracy in both mapping and tracking. A loosely-coupled system provides flexibility to the combination of sensors with lower requirement of timestamp synchronization, and typically requires lower computational cost.

In recent years, as event camera becoming available, event-based SLAM [18] has also emerged. Since event cameras output low-latency, asynchronous, faster and higher dynamic range visual measurements, event-based SLAM is becoming a hot research topic.

*C. SLAM Systems and Difficulties*

A key component for SLAM system development is the hardware. A typical setup for SLAM R&D in a research institute is using ROS (Robot Operating System) [19] as the runtime development environment on a PC, which is fragmented and hard to conduct embedded runtime performance evaluation. There are several hardware setups for SLAM where datasets are publicly available [12, 20, 21] to evaluate SLAM approaches. However, these setups are not commercially available.

There are several embedded developer kits available (e.g., Samsung SoC based ODROID [22], Nvidia SoC based TX1/TX2 [23, 24], Qualcomm SoC based Intrinsyc [25]), but it takes an expert in firmware and driver to connect the platform with different sensors, which is time-consuming and not of the interest of most SLAM researchers. Currently, there are several available commercial developer products for SLAM. Intel Euclid [26] integrates of multiple sensors and the Intel Atom Cores, but with no ARM support. Google Tango [27] is limited to monocular camera. Microsoft HoloLens [28] has limitations of HPU access, which is not friendly as a computing platform, and Windows only platform. Apple ARKit [29] is limited to Apple hardware and iOS platform. The recent Google ARCore [30] has the potential to enable SLAM on hundreds of millions of Android devices but again, researchers will be limited by the form factor of the device.

A truly "SLAM-ready" hardware must be capable of providing: 1) accurate timestamp for sensors via hardware synchronization; 2) accurate calibration in a massive scale, and 3) an easy-to-use PC and embedded platform. The Skybotix VI-Sensor [31] is one of the first visual inertial sensor modules, but currently it's discontinued after the company got acquired by GoPro. We PerceptIn released the Visual Inertial Module (product name: Old Ironsides) [32] in February 2017 for the purpose of providing a "SLAM-ready" sensor module. The module has been widely adopted by 200+ researchers and company developers. However, as a sensor module, it has to be connected to some computing system, such as PC or Nvidia developer board, to run SLAM algorithms.

To sum up, for an optimal design of a SLAM system to truly work in real-world applications, the hardware setup and the SLAM algorithms must be collaboratively designed.

*D. Our Contributions*

We present the design and evaluation of the PerceptIn Robotics Vision System (PIRVS). Our contributions are summarized in two folds:

- A novel all-in-one multi-sensor computing hardware for SLAM as shown in Fig. 1. It is equipped with sensors specifically for visual-inertial SLAM, including a carefully crafted stereo camera and an inertial measurement unit (IMU), and a low-power ARM SoC for computation. The sensors are accurately synchronized and calibrated.

- A visual-inertial SLAM approach which features a novel and flexible sensor fusion algorithm. Our algorithm includes tightly-coupled visual-inertial tracking and parallel mapping, and is capable to loosely couple with additional sensor modalities for different applications.

Our system achieves similar performance compared to other state-of-the-art systems, and is proven to work efficiently on our hardware. Our system is designed to be modularized and easy to interact with different applications via advanced loosely coupled sensor fusion.

The rest of the paper is structured as follows. In Section II, we introduce the details of our hardware design and implementation. In Section III, we present the software stack of our system. In Section IV, we present our SLAM architecture including image processing front-end, tightly-coupled visual-inertial tracking, parallel mapping, and finally the loosely-coupled sensor fusion for integrating additional sensor modalities for specific applications with our system. In Section V, we introduce detailed comparison with other state-of-the-art visual-inertial SLAM systems using public datasets. Finally, we present our conclusion in Section VI.

## II. PIRVS HARDWARE

PIRVS device (product name: Ironsides) has two OmniVision global shutter CMOS image sensors capturing 640×480 resolution color images at 60 fps. The lenses are fixed-focal fisheye lenses with 133° horizontal field of view (HFOV) and 100° vertical field of view (VFOV). The baseline between the image sensors is 65 mm. The device also has a commercial grade InvenSense IMU reporting inertial measurements at 200 fps. The device is equipped with a dual-core ARM Cortex A72 and quad-core ARM Cortex A53 SoC, which also has an ARM Mali-T860MP4 GPU embedded with four shader cores with shared hierarchical tiler. The device has a 2GB RAM and 8GB internal storage. Furthermore, it provides two interfaces for external communication: 1) a 3-pin GPIO interface to connect with any robotics device, (e.g., ground robot, drone), and 2) standard USB 3.0 which also enables itself functioning as a "plug-and-play" sensor.

Figure 2.  PIRVS Hardware Timestamp Synchronization.

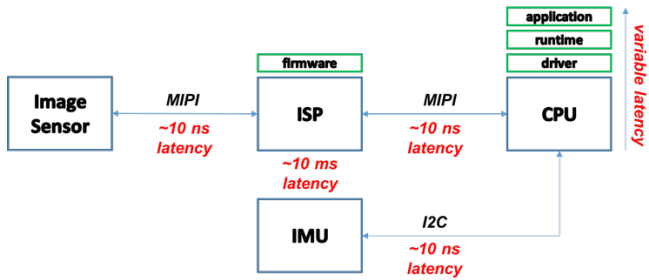

On PIRVS hardware, we use ISP to trigger hardware synchronization between the stereo cameras and to trigger auto-exposure and image pre-processing to improve image quality. As shown in Fig. 2, the latencies between CMOS image sensor and ISP, and ISP and CPU, are both roughly 10ns through the MIPI interface. This latency number also holds for the IMU data which is triggered through the I2C interface. The image pre-processing on ISP takes ~10ms which is still reasonable especially the latency is consistent with little uncertainty. We choose not to timestamp the images at ISP level because the image timestamp needs to be synchronized with the IMU timestamp with the same clock given the synchronization between clocks is quite complicated. Therefore, we timestamp both image and IMU data at the driver level of the host CPU which is tolerable while not over-complicating the hardware design. Furthermore, all of the above operations are done in native driver level without experiencing the latencies and instabilities from the runtime and from the application level, so the timestamp ordering of IMU and camera readings is guaranteed to be correct.

We designed a factory calibration pipeline to determine the intrinsics and extrinsics of the stereo camera, IMU intrinsics (skew, scale, and biases) and extrinsics, and additional temporal offset between IMU and the stereo camera for each device in a massive scale. Our temporal calibration is modified based on Kalibr [33]. So far, we reached an average of 0.2 pixel re-projection error for camera intrinsics and 0.7 pixel for camera extrinsics. We measured an average of 17ms timestamp offset with 2ms variance between the stereo camera and IMU. The entire calibration process of takes 3 minutes per device.

## III.  PIRVS System

On top of the PIRVS hardware, we design the PIRVS system running an Android system without middleware layer which means: 1) it supports native C/C++ code through Android NDK, and 2) it does not have the typical Android overhead (e.g., garbage collection from Java layer). The dependencies have been pre-installed in the image. Our SDK is cross-compiled on Linux. The potential acceleration via ARM NEON instrument sets is fully available for the users to take advantage of the heterogeneous ARM-based system. Fig. 3 illustrates our system architecture. In the native layer, in contrast to ROS, our design ensures that sensor data acquisition and algorithm execution are in a real-time fashion with minimum overhead. Therefore, a main thread is created to invoke each module in the pipeline at a rate which is fast enough for the most frequent data (e.g., IMU data). The modules which require intensive computation runs in their own threads to avoid blocking the main thread.

Figure 3.  PIRVS System Architecture.

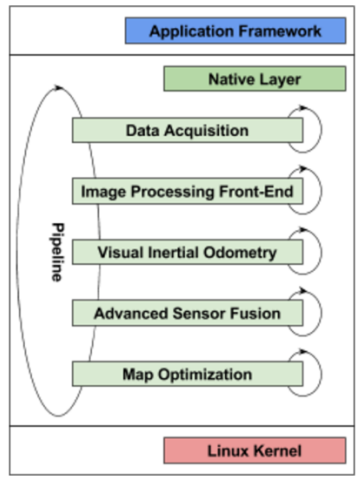

## IV.  PIRVS Algorithm

We implemented a visual-inertial tracking algorithm localizing the device within a 3D map that is simultaneously being built in a separate thread. There are three major components in our algorithm: 1) highly optimized image processing front-end extracting image features and matching them with the 3D map; 2) EKF-based tightly-coupled visual-inertial tracking with IMU propagation and state update with 3D-2D feature correspondences; and 3) mapping with prior poses from the tracking thread. Our algorithm can be loosely-coupled with other external sensor, such as wheel odometry, GNSS (Global Navigation Satellite System), for specific applications.

### A.  Image Processing Front-End

We developed a pipeline to efficiently detect corner features spreading across the entire images. Upon receiving an image with all the pre-processing from the ISP on the PIRVS hardware, we perform histogram equalization to adjust contrast to allow more features being detected to handle environments with low illumination, as shown in Fig. 4. Then, we use optical flow [34] to track the features from the previous image to the current image. If the number of tracked features is low, we detect new features in regions where tracked features are not available. We select the features through spatial binning to ensure that the features are spread uniformly across the entire field-of-view. Finally, we calculate a binary descriptor for each feature.

Figure 4.  An Example Dark Environment Image Before and After Histogram Equalization.

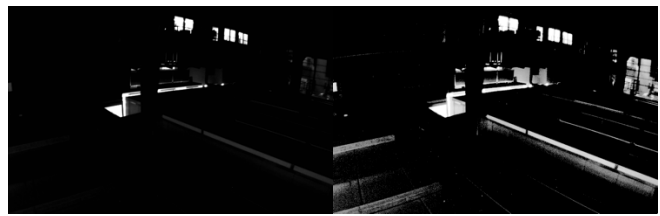

Once the features are detected, we establish the association of each feature to a 3D map point in the map. Details of the map representation are presented in Section C. A feature obtained through optical flow will have the same 3D map point association of that feature from the previous image. For the rest of the features, we find the best matching 3D map point for each feature. A 2D feature and a 3D map point is a match if the re-projection error given the current tracking pose is within a threshold. The match is ranked by the distance between the descriptors of the 2D feature and the 3D map point. The smaller the distance, the better the match is.

Since both our feature detection and matching pipelines make full use of temporal information, our image processing front-end is very efficient as we do not repeatedly detect new features in the entire image for each frame. We leave the more time-consuming stereo feature matching to the mapping thread and perform it only when necessary.

### B. Tightly Coupled Visual-Inertial Tracking

Our tracking algorithm uses the iterated Extended Kalman Filter (EKF) framework. Our filter state at time $t$ is parameterized as $X_t = [q_G^{B_t} \ p_{B_t}^G \ v_{B_t}^G \ b_a \ b_g]$, where $q_G^{B_t}$ is the unit quaternion describing the rotation from the global frame $\{G\}$ to the IMU body frame $\{B\}$, $p_{B_t}^G$ and $v_{B_t}^G$ are the IMU position and velocity with respect to $\{G\}$, and finally $b_a$ and $b_g$ are gyroscope and accelerometer biases.

With our well-chosen sensors and accurate hardware synchronization, we are able to precisely calibrate the IMU-camera extrinsics offline, allowing us to exclude online extrinsics calibration. This not only speeds up the computation at run-time but also provides better convergence and accuracy in tracking.

Whenever a new IMU measurement (angular velocity $w_t$ and linear acceleration $a_t$) is received, we propagate the state by the kinematic equations as follows:

$$\dot{R}_G^{B_t} = \left(R_{B_t}^G\right)^T = \left(R_{B_{t-1}}^G R_{B_t}^{B_{t-1}}\right)^T \quad (1)$$

$$R_{B_t}^{B_{t-1}} = f\left((w_t - b_g)\Delta t\right) \quad (2)$$

$$\Delta v_{B_t} = \left(R_{B_t}^G(a_t - b_a) + g^G\right)\Delta t \quad (3)$$

$$\dot{v}_{B_t} = v_{B_{t-1}} + \Delta v_{B_t} \quad (4)$$

$$\dot{p}_{B_t} = p_{B_{t-1}} + v_{B_{t-1}}\Delta t + \frac{1}{2}\Delta v_{B_t}\Delta t \quad (5)$$

where $f(\cdot)$ is the function converting axis-angle rotation to rotation matrix, $\Delta v_{B_t}$ is the velocity increment, and $g^G$ represents the gravity correction in the global coordinate system. EKF update happens when measurements (a set of correspondences between 2D image features and 3D map points) from an image is received. Residual for each correspondence is the re-projection error of each 3D map point $P^G$ in the global coordinate,

$$\begin{bmatrix} u \\ v \end{bmatrix} = \Pi\left(R_B^{C_i}\left(\dot{R}_G^{B_t}\left(P^G - \dot{p}_{B_t}^G\right)\right) + p_B^{C_i}\right) \quad (6)$$

where ($R_B^{C_i}, p_B^{C_i}$) is the transform from the IMU to the coordinate of the $i$-th camera obtained through offline calibration, and $R_G^{B_t}$ is the rotation matrix of the quaternion $q_G^{B_t}$. $\Pi$ is the perspective projection of a 3D point to the image coordinate $\begin{bmatrix} u \\ v \end{bmatrix}$. The actual EKF residual is constructed by concatenating residuals from all the measurements.

### C. Parallel Mapping

We maintain a 3D sparse map in the global coordinate parallel to our visual-inertial tracking. The map consists of a set of 3D map points and a set of keyframes with constraints between the 2D observed features from the stereo camera and the 3D map points.

A keyframe is inserted when the device sees a portion of the environment that is not covered by the map. Thus, we compare the tracking pose ($q_G^{B_t}, p_{B_t}^G$) and the poses of the existing keyframes. A tracking pose becomes a keyframe if the smallest heading difference between the device and the existing keyframes is larger than an angle, or if the distance to any keyframes is larger than a threshold. Once a keyframe is selected, we add the set of correspondences between 2D features and the existing 3D map points, obtained from our image processing front-end, to the map.

Upon inserting a new keyframe, new map points are created from both spatial and temporal correspondences from 2D features in the keyframes that are not yet being associated to the 3D map. Temporal correspondences are obtained through our image-processing front-end, and spatial correspondences are established by matching features from the stereo images. Note, with our accurately calibrated stereo camera, we can efficiently search for spatial correspondences using epipolar constraint. Finally, we triangulate each set of corresponding 2D features into a 3D map point.

The 6-dof pose of the keyframes and the 3D locations of the map points are refined through a standard bundle adjustment approach [35]. For efficiency, only a portion of the map is being refined at a time. At each optimization, we select a subset of the keyframes to be adjusted. This subset includes the latest keyframe and the frames with sufficient shared field-of-view. The map points and keyframe poses within the shared field-of-view are being refined.

The map is being cleaned up after each optimization to ensure the quality of the map and to keep the size of the map tractable. We remove constraints with large re-projection errors. A map point or a keyframe without sufficient number of constraints are also being removed from the map. In addition, if the number of keyframes reaches a limit, older keyframes are being removed.

### D. Loosely Coupled Sensor Fusion

To truly utilize a SLAM system for a specific application, we need to integrate the tracking pose and the map from SLAM with other sensors available for that application. For an indoor ground-traveling robot, we can enhance the robustness of tracking by combining the SLAM system with wheel odometry. For outdoor applications, like autonomous drones or vehicles, we can integrate GNSS (Global Navigation Satellite System) with our SLAM system to provide localization in the absolute map. The modularized design of PIRVS makes it easy to satisfy different demands. In this section, an example of integrating PIRVS system with GNSS system is presented. The systems are loosely coupled to achieve better performance and to increase the overall robustness. The term "loosely coupled" here is defined as integrating the two systems using higher level information (position, velocity, and orientation), which are provided by PIRVS system and GNSS independently, and then fused by a centralized EKF based fusion algorithm, as shown in Fig. 5. Calibration for the transformation between SLAM and GNSS is done offline by using the first pose estimation to define both origins and then solving the orientation alignment known as Wahba's problem [36].

Figure 5. A PIRVS Loosely Coupled Sensor Fusion Example.

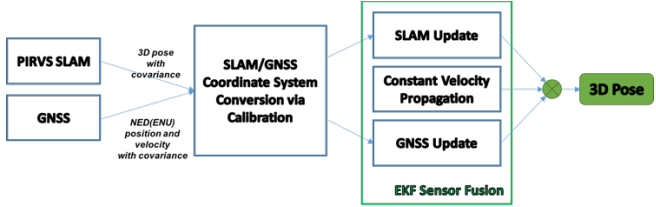

We use $[p\ v\ \psi]$ to present the 9x1 vector containing position, velocity, and orientation (axis angle representation). Eq. 7-9 represent constant velocity propagation, SLAM update, and GNSS update respectively, where $\mu$ is the zero mean Gaussian random noise with covariance, $\gamma$ is the zero mean Gaussian measurement noise with covariance given both SLAM and GNSS updates are treated as measurement observations by the sensor fusion module. $R_{GS}^B$ represents the rotation matrix from GNSS to SLAM's IMU body frame.

$$\begin{bmatrix} p \\ v \\ \psi \end{bmatrix}_{t+1} = \begin{bmatrix} I_3 & dt*I_3 & 0_3 \\ 0_3 & I_3 & 0_3 \\ 0_3 & 0_3 & I_3 \end{bmatrix} \begin{bmatrix} p \\ v \\ \psi \end{bmatrix}_t + \begin{bmatrix} \mu_p \\ \mu_v \\ \mu_\psi \end{bmatrix}_t \quad (7)$$

$$\begin{bmatrix} p \\ v \\ \psi \end{bmatrix}_t = \begin{bmatrix} R_{GS}^B & 0_3 & 0_3 \\ 0_3 & R_{GS}^B & 0_3 \\ 0_3 & 0_3 & R_{GS}^B \end{bmatrix} \begin{bmatrix} I_3 & 0_3 & 0_3 \\ 0_3 & I_3 & 0_3 \\ 0_3 & 0_3 & I_3 \end{bmatrix} \begin{bmatrix} p \\ v \\ \psi \end{bmatrix}_t + \begin{bmatrix} \gamma_p \\ \gamma_v \\ \gamma_\psi \end{bmatrix}_t \quad (8)$$

$$\begin{bmatrix} p \\ v \end{bmatrix}_t = \begin{bmatrix} I_3 & 0_3 & 0_3 \\ 0_3 & I_3 & 0_3 \end{bmatrix} \begin{bmatrix} p \\ v \\ \psi \end{bmatrix}_t + \begin{bmatrix} \gamma_p \\ \gamma_v \end{bmatrix}_t \quad (9)$$

## V. EXPERIMENTAL RESULTS

We conducted a few experiments to show that our PIRVS software approach achieves comparable performance (accuracy and speed-wise) compared to other state-of-the-art visual-inertial SLAM approaches on PC using public datasets while being considerably more efficient on the PIRVS hardware.

### A. Datasets, Comparison Approaches, and Metrics

There are several high-quality visual inertial datasets [12, 20, 21]. We choose to evaluate on PennCOSYVIO [20] and EuRoC [21] datasets for the following reasons. First, these datasets provide ground truth trajectories. Second, the frame rate of IMU from these datasets is at least 100 fps, and the frame rate for stereo images is at least 10 fps. Third, the camera lenses from these datasets are not fisheye lens. The reason for excluding fisheye camera datasets is that several state-of-the-art approaches haven't properly leveraged wide FOV image measurements due to the severe distortion. Finally, both datasets covered challenging scenarios. EuRoC is an indoor dataset with dramatic fast motion and challenging lighting conditions. PennCOSYVIO has a long outdoor path as well as an outdoor/indoor transition in an environment with extended glass walls on both outside and inside and with many repetitive patterns.

We choose to compare to OKVIS [13] and VINS-MONO [14], which are two representative state-of-the-art tightly-coupled visual-inertial SLAM algorithms. They have both open-source implementations and configuration parameters which are essential for fair comparison. We exclude loosely-coupled approaches (e.g., [11]), as they cannot fully utilize IMU and image data, resulting in information loss [16]. We also exclude direct methods and event-camera based approaches because direct methods typically incur higher computational cost making them not suitable for existing embedded resource-constrained platforms, and event-camera based approaches require event cameras which are currently not mature for mass production and deployment. We want to specially note that we believe that the University of Minnesota MARS Lab's VINS [12] system with high quality dataset is one of the state-of-the-art approaches desirable for comparison. However, due to the lack of open-source implementation, we had to exclude it from our comparison. For the same reason, we are not able to compare to MSCKF [15] or MSCKF 2.0 [16], which are representative tightly-coupled filtering-based VIO approaches using sparse features for update.

For metrics, we choose to use Absolute Trajectory Error (ATE) and Relative Pose Error (RPE) [37]. ATE measures the accuracy of the entire trajectory so drifts or rotational error will be accumulated. RPE measures the errors between relative poses in every time period, which shows how well a system estimates the movement of the device at a given time period. We utilize the TUM RGB-D SLAM evaluation toolset [37] to align the coordinate of a SLAM system to the ground-truth's coordinate via the closed-form Horn's Method [38] so that the trajectories can be compared.

### B. Accuracy

We evaluated the accuracy of OKVIS, VINS-MONO, and our proposed PIRVS system on the AS dataset from PennCOSYVIO [20] and the MH_04_Difficult dataset from EuRoC [21] using a 64-bit Ubuntu laptop with Intel Core i7-6560U CPU @ 2.20GHz x 4 and 16GB memory. Fig. 6 shows the ATE errors along the trajectories. The top row is the result from OKVIS, and the second row is from VINS-MONO. Our PIRVS is shown at the bottom row. Table I shows the complete statistics of the accuracy evaluation. Green means the best while red means the worst.

All three approaches produced trajectories for the entire sequence on both datasets. Our PIRVS approach consistently performs the best based on ATE RSME and STD metrics. OKVIS has the largest trajectory error in the return trip of EuRoC dataset possibly due to error accumulation in its local mapping. VINS-MONO performs the worst in the PennCOSYVIO dataset due to poor loop closure performance caused by the many repetitive features in the ~150m indoor/outdoor transition path. Note that in contrast to VINS-MONO, we do not do explicit loop closure. Instead, we keep a local map similar to OKVIS. There are two reasons for us to outperform both approaches on absolute trajectory error. First, our tightly-coupled filter design for tracking has less dependence on the map, allowing us to continue tracking even in challenging environments when correspondences are hard to find. Our tracking algorithm can update the pose estimate with as few as two or even one correspondence, while VINS-MONO and OKVIS require at least three correspondences for non-ambiguous pose estimation [39]. Second, our mapping algorithm is more accurate because we carefully add high-quality 2D-3D constraints and constantly remove poor constraints in our optimization. This helps us to achieve good tracking performance and to avoid drifting heavily in challenging scenarios. This also make us different from MSCKF [15, 16] or SR-ISWF [17], as they do not perform separate mapping.

Figure 6. ATE/RPE Comparisons With Default Settings.

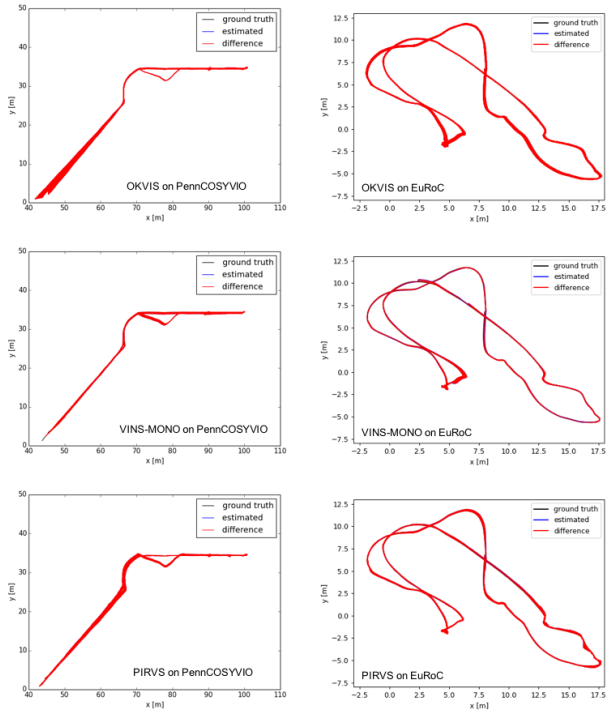

TABLE I. ATE/RPE ERROR COMPARISON

|  | EuRoC_MH04_Difficult Absolute Trajectory Error | EuRoC_MH04_Difficult Relative Pose Error | PennCOSYVIO_AS Absolute Trajectory Error | PennCOSYVIO_AS Relative Pose Error |
|---|---|---|---|---|
| OKVIS | RMSE: 0.184569 m STD: 0.056455 m | RSME_T: 0.280032 m STD_T: 0.135998 m RSME_R: 1.008995 deg STD_R: 0.528699 deg | RMSE: 0.629347 m STD: 0.353461 m | RSME_T: 1.683979 m STD_T: 1.239022 m RSME_R: 3.096804 deg STD_R: 2.279090 deg |
| VINS-MONO | RMSE: 0.139038 m STD: 0.088764 m | RSME_T: 0.271694 m STD_T: 0.148698 m RSME_R: 1.535694 deg STD_R: 1.104931 deg | RMSE: 1.276582 m STD: 0.680134 m | RSME_T: 1.985967 m STD_T: 1.344148 m RSME_R: 2.121836 deg STD_R: 1.020091 deg |
| PIRVS | RMSE: 0.122709 m STD: 0.049498 m | RSME_T: 0.339272 m STD_T: 0.185003 m RSME_R: 1.450916 deg STD_R: 0.733636 deg | RMSE: 0.592735 m STD: 0.180257 m | RSME_T: 1.600193 m STD_T: 1.050176 m RSME_R: 2.654375 deg STD_R: 1.215665 deg |

However, our approach never performs the best on the rotational part of RPE. To improve this, as a future work, we plan to add relative rotation constraints from gyro integration to the bundle adjustment, similar to VINS-MONO and OKVIS, which could improve the map quality in cases of fast rotations. We also notice that our approach performs worse than VINS-MONO and OKVIS in low lighting conditions, even with the use of histogram equalization technique. We also plan to improve this in the future work.

*C. Speed*

We performed several key speedups for the full pipeline on PIRVS hardware. For our image-processing front-end, NEON optimization has been used for image pre-processing, feature detection, and descriptor calculation. We compute the above image processing for each image in the stereo camera in parallel. The complete image processing front-end takes 30ms on an embedded system. Similar performance is only seen on the University of Minnesota MARS Lab's Nvidia TX1 based hardware and software system [12]. For mapping, we create a thread specifically for bundle adjustment to prevent the main thread being blocked by optimization.

We perform an extensive timing profile comparison between PIRVS and OKVIS on the AS dataset from PennCOSYVIO. Note, we excluded VINS-MONO in the comparison because its implementation only runs on ROS, which is hard to be ported onto Android embedded systems. The comparison is shown in Table II. The end-to-end timing of our algorithm on PIRVS hardware is ~10 fps, in comparison with OKVIS ~8 fps. For our system, the timing for mapping is the averaged time spent on optimization processed in a separate thread (e.g., the total processing time of the bundle adjustment thread divided by the total number of images processed). Thus, at run-time, our system is ~23 fps. For OKVIS, the timing for tracking and mapping is the time spent for optimization and marginalization. Note that with our limited efforts of optimizing OKVIS on the PIRVS hardware, we notice some threading synchronization which causes latency from 20ms up to 200ms especially on the EuRoC dataset.

TABLE II. TIMING COMPARISON ON PIRVS HARDWARE

|  | OKVIS on PC | OKVIS on PIRVS hardware | PIRVS on PC | PIRVS on PIRVS hardware |
|---|---|---|---|---|
| Image Processing | 5.4 ms | 16.3 ms | 8.7 ms | 30 ms |
| Matching | 4.6 ms | 19.3 ms | 0.7 ms | 3.8 ms |
| Tracking | 39.7 ms | 93.8 ms | 3.6 ms | 8.5 ms |
| Mapping |  |  | 5.9 ms | 57.2 ms |
| End-To-End | 49.7 ms | 129.4 ms | 18.9 ms | 99.5 ms |

VI. CONCLUSIONS

In this paper, we presented the PerceptIn Robotics Vision System (PIRVS) system for visual-inertial SLAM applications. We presented the details of the PIRVS hardware design including sensors, computing resource and the design of our high efficiency and low power system architecture. We also presented the PIRVS algorithms including a highly-optimized image processing pipeline, tightly-coupled visual-inertial tracking, parallel mapping design, and advanced sensor fusion with additional sensor modalities. Detailed experimental results are presented including accuracy comparison, timing measurements, and key factors of performance (e.g., implementation details on image processing and map optimization) are discussed and validated. The results demonstrate that our algorithm reaches state-of-the-art visual-inertial SLAM performance, and that our PIRVS system, the co-design of the proposed hardware and our algorithm, is efficient, powerful, and user friendly. Our PIRVS system is available upon the purchase of PIRVS hardware [40].

Our future work includes three directions. First, we will continuously improve the hardware design. Second, we plan to implement concrete applications integrating external sensors into the PIRVS system. Third, we plan to improve our SLAM algorithm by using sliding window of poses in tracking and by enhancing our image-processing pipeline for environments with challenging illumination condition.

ACKNOWLEDGMENT

Thanks Yen-Cheng Liu for calibrating the PIRVS hardware. Thanks Qingyu Chen and Weikai Li for developing and testing algorithms and applications on PIRVS software. Thanks Chandrahas Jagadish Ramalad for collecting data and evaluating different SLAM systems.